\documentclass[sn-mathphys-num]{sn-jnl}

\usepackage{graphicx}%
\usepackage{amsmath,amssymb,amsfonts}%
\usepackage{amsthm}%
\usepackage{mathrsfs}%
\usepackage[title]{appendix}%
\usepackage{textcomp}%
\usepackage{manyfoot}%
\usepackage{algorithm}%
\usepackage{algorithmicx}%
\usepackage{algpseudocode}%
\usepackage{listings}%

\usepackage{epsfig} 
\usepackage{mathptmx} 
\usepackage{times} 
\usepackage{amsmath} 
\usepackage{amsmath,array}
\usepackage{amssymb}  
\usepackage{subfigure}
\usepackage{subcaption}
\usepackage{hyperref}

\usepackage[export]{adjustbox}
\usepackage{siunitx}
\usepackage{makecell}
\usepackage{indentfirst}
\usepackage{verbatim}
\usepackage{diagbox}
\usepackage{tabularx}
\usepackage{multirow}%
\usepackage{booktabs}
\usepackage{float}
\usepackage{tabularray}

\begin{document}

\title[Shoulder Range of Motion Rehabilitation Robot Incorporating Scapulohumeral Rhythm for Frozen Shoulder]{Shoulder Range of Motion Rehabilitation Robot Incorporating Scapulohumeral Rhythm for Frozen Shoulder}


\author[1,2]{\fnm{Hyunbum} \sur{Cho}}
\author[1,2]{\fnm{Sungmoon} \sur{Hur}}
\author[3]{\fnm{Joowan} \sur{Kim}}\email{joowan@snu.ac.kr}
\equalcont{These authors contributed equally to this work.}
\author[3]{\fnm{Keewon} \sur{Kim}}
\author[1,2,4]{\fnm{Jaeheung} \sur{Park}}\email{park73@snu.ac.kr}
\equalcont{Joowan Kim and Jaeheung Park are the corresponding authors.}

\affil[1]{\orgdiv{Department of Intelligence
and Information}, \orgname{Seoul National University}, \orgaddress{\city{Seoul}, \postcode{08826}, \country{Republic of Korea}}}

\affil[2]{\orgname{Blue Robin inc.}, \orgaddress{\city{Seoul}, \postcode{06524}, \country{Republic of Korea}}}

\affil[3]{\orgdiv{Department of Rehabilitation Medicine}, \orgname{Seoul National University Hospital}, \orgaddress{\city{Seoul}, \postcode{03080},\country{Republic of Korea}}}

\affil[4]{\orgdiv{ASRI, RICS}, \orgname{Seoul National University and Advanced Institutes of
Convergence Technology}, \orgaddress{\city{Seoul}, \postcode{08826}, \country{Republic of Korea}}}


\abstract{This paper presents a novel rehabilitation robot designed to address the challenges of Passive Range of Motion (PROM) exercises for frozen shoulder patients by integrating advanced scapulohumeral rhythm stabilization. Frozen shoulder is characterized by limited glenohumeral motion and disrupted scapulohumeral rhythm, with therapist-assisted interventions being highly effective for restoring normal shoulder function. While existing robotic solutions replicate natural shoulder biomechanics, they lack the ability to stabilize compensatory movements, such as shoulder shrugging, which are critical for effective rehabilitation. Our proposed device features a 6 Degrees of Freedom (DoF) mechanism, including 5 DoF for shoulder motion and an innovative 1 DoF \textit{Joint press} for scapular stabilization. The robot employs a personalized two-phase operation: recording normal shoulder movement patterns from the unaffected side and applying them to guide the affected side. Experimental results demonstrated the robot’s ability to replicate recorded motion patterns with high precision, with Root Mean Square Error (RMSE) values consistently below 1 degree. In simulated frozen shoulder conditions, the robot effectively suppressed scapular elevation, delaying the onset of compensatory movements and guiding the affected shoulder to move more closely in alignment with normal shoulder motion, particularly during arm elevation movements such as abduction and flexion. These findings confirm the robot’s potential as a rehabilitation tool capable of automating PROM exercises while correcting compensatory movements. The system provides a foundation for advanced, personalized rehabilitation for patients with frozen shoulders.}





\maketitle
\section{Introduction}\label{sec1}
\label{sec:introduction}

Frozen shoulder, also known as adhesive capsulitis, is a debilitating condition characterized by pain and progressive loss of shoulder movement \cite{c1}. The underlying cause of a frozen shoulder is not fully understood, and it is associated with inflammation and thickening of the capsule that surrounds the shoulder joint where the glenoid of the scapula and the proximal humerus meet \cite{c1, c2}. The condition affects approximately 2-5\% of the general population, with a higher incidence in people aged in the mid-50s \cite{c3}. 

Frozen shoulder problems extend beyond limited Range of Motion (ROM), involving compensatory movements that can lead to secondary issues. These compensations mainly include excessive scapular upward rotation and trunk adjustments during arm elevation \cite{c4}. Especially, excessive scapular upward rotation results as shoulder shrugging, which directly affects the scapulohumeral rhythm \cite{c9, c10}.

Scapulohumeral rhythm, traditionally described as a 2:1 ratio between glenohumeral elevation and scapular upward rotation, but in reality it is a nonlinear motion, is crucial for shoulder stability \cite{c5, c6, c8}. Its disruption due to compensatory movements can develop secondary problems, such as impingement syndrome, rotator cuff dysfunction, and shoulder instability \cite{c9, c11}. Moreover, this disruption can delay the recovery of shoulder injuries and increase the likelihood of the condition becoming chronic \cite{c38}.

Exercise is the most reliable and versatile treatment for frozen shoulder, demonstrating effectiveness across all stages of the condition, including post-operative rehabilitation. Notably, it also shows value in addressing scapular dyskinesis \cite{c2,c3,c4}. While the treatment for frozen shoulder can be divided into conservative and operative treatment, conservative approaches can resolve approximately 90\% of early-stage cases. The conservative treatment includes medications, physical therapy, exercise, steroid injections, and hydrodilation \cite{c12,c13}. Among these, exercise distinguishes itself by comprehensively addressing both mobility deficits and scapular dysfunction.

Therapist-assisted Passive ROM (PROM) exercises are an effective frozen shoulder treatment, as it suppresses compensatory movements and facilitates normal movement patterns \cite{c42}. This approach is superior to self PROM exercises, which could often leads to compensatory movements \cite{c3,c12,c41}. In assisted PROM exercises, the therapist manually controls the patient's shoulders and arms, using visual and tactile feedback to inhibit compensatory behaviors. One of the primary methods for suppressing compensatory movements is stabilizing the scapula. This stabilization allows for proper scapulohumeral rhythm and ensures that movement occurs within a normal ROM, safely progressing while minimizing injury risk. Importantly, this technique promotes active and conscious control of scapular muscles, re-engaging correct muscle activation patterns and improving overall shoulder function \cite{c42, c43, c16}.

Since therapist-assisted ROM exercise requires specialized expertise, which is not always available, a rehabilitation robot capable of performing assisted PROM exercise while correcting compensatory movements could improve frozen shoulder treatment. This would enhance therapy efficiency by automating repetitive processes and consistently correcting patient's movements. Also by reducing healthcare professionals' workload and collecting valuable data for patient evaluation, such a robot could optimize treatment strategies.

Numerous shoulder and upper-limb rehabilitation robots are currently developed or commercially available. End-effector type robots are primarily designed to facilitate repetitive training of daily activities or motions, with a focus on overall upper-limb muscle function and neurological recovery. These robots are particularly suited for patients with upper-limb motor impairments following a stroke \cite{c39, c40, c45, c46, c47}. However, they generally do not specifically target the shoulder joint, limiting their ability to stabilize the scapula and perform precise ROM exercises tailored for frozen shoulder patients. 

In contrast, exoskeleton or wearable-type rehabilitation robots are often designed to assign robotic joints directly to corresponding human motor functions, including the shoulder, for upper-limb rehabilitation \cite{c48, c49, c50, c51, c52, c53, c54, c55, c20}. Some of these robots incorporate degrees of freedom for shoulder girdle movements, such as scapular motion, to align the robotic axis with the shoulder joint during rotation \cite{c50, c51, c52, c53, c54, c55, c20}. While these designs consider mimicking scapular motion for axis alignment, they are primarily targeted at stroke patients, focusing on addressing misalignment between the robot and the glenohumeral joint to enable natural motion and prevent injuries. However, these robots do not directly stabilize the scapula, resulting in limited capability to effectively control compensatory movements.

Additionally, specific rehabilitation robots have been developed for frozen shoulder treatment, including the NTUH-II and SRR \cite{c17, c18, c19}. The NTUH-II is a robotic arm designed for shoulder rehabilitation, supporting both passive and active robot-assisted training modes while incorporating functionality to identify the patient’s ROM limits, making it suitable for addressing the needs of frozen shoulder patients \cite{c17, c18}. Similarly, the SRR is a rehabilitation robot designed for frozen shoulders, featuring both passive and active exercise modes and utilizing motion capture data to guide passive training \cite{c19}. However, despite their targeted designs, these robots lack the capability to directly stabilize the scapula, thereby limiting their effectiveness in maintaining the patient’s normal scapulohumeral rhythm and suppressing compensatory movements during rehabilitation exercises.

\begin{figure}[t]
      \centering
      \includegraphics{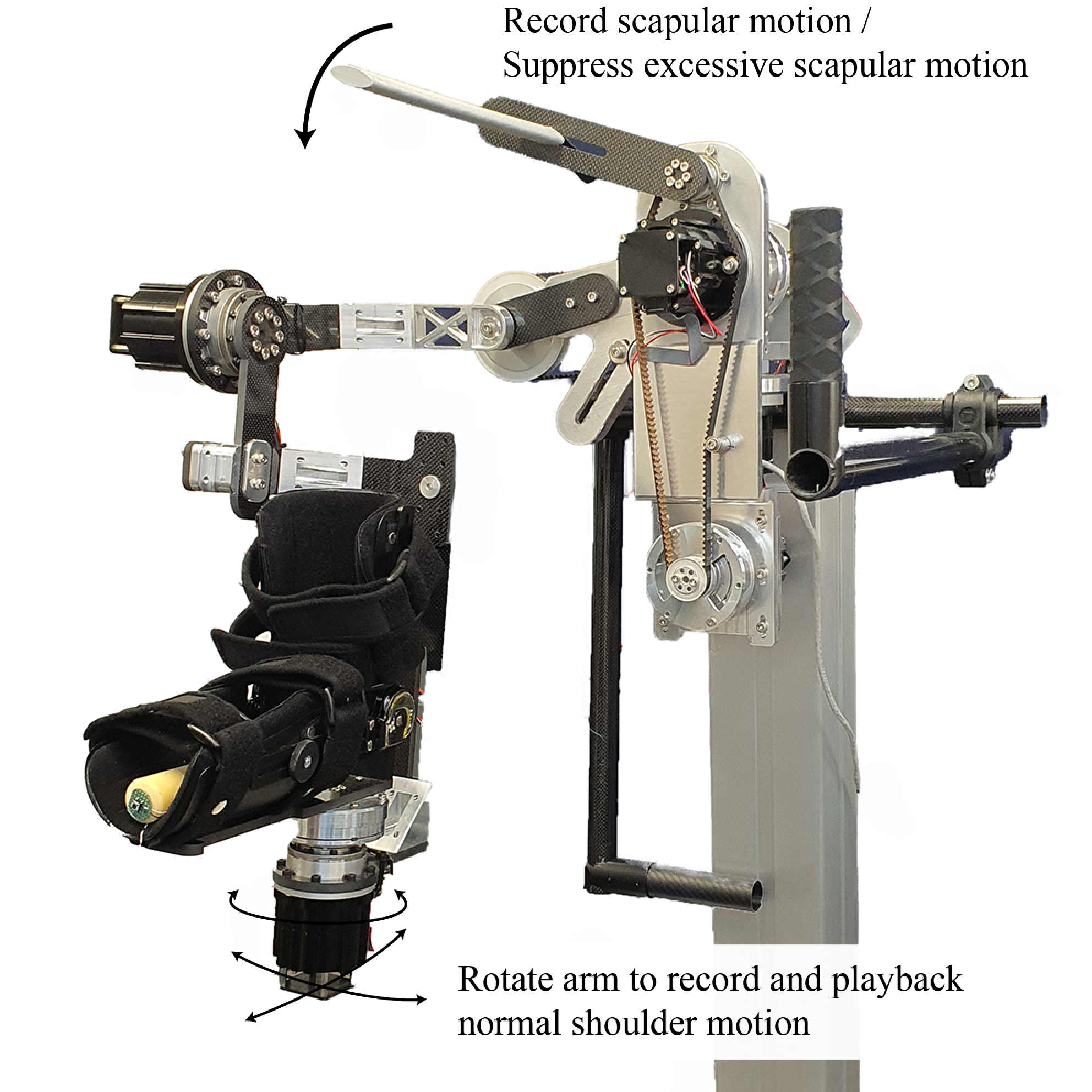}
      \caption{Proposed rehabilitation robot}
      \label{fig:prototype}
   \end{figure}

This paper presents a novel rehabilitation robot for frozen shoulder which is designed to correct compensatory movements while considering the patient's normal scapulohumeral rhythm, as shown in Fig. \ref{fig:prototype}. The robot features several key components, including:

\begin{itemize}
\item A ROM exercise mechanism with 2 Degree of Freedom (DoF) for translation and 3 DoF for rotation to replicate natural shoulder motion, including scapulohumeral rhythm. This mechanism records natural, normal shoulder motion as a reference for ROM exercises and plays back the recorded reference to guide ROM exercise.
\item A novel 1 DoF mechanism is integrated into the robot to measure and control scapular motion, specifically to stabilize the scapula by suppressing shoulder shrugging. This mechanism records shoulder elevation from a position above the acromion, allowing it to estimate scapular movement and capture the normal scapulohumeral rhythm. Additionally, it guides exercises by suppressing compensatory movements, ensuring that patients follow the recorded normal motion pattern effectively.
\item Structures that limit the trunk's compensatory movements such as rotating or leaning.
\item The rehabilitation protocol records unaffected arm motion patterns for personalized playback treatment on the affected side.
\end{itemize}

The paper is structured as follows: Section II provides a detailed description of the decision-making process to determine the mechanism of the robot, including its unique shoulder-pressing and ROM exercise mechanisms. Section III explores the controls and algorithms. Section IV evaluates the robot’s overall performance through various experimental results and their analysis. Section V presents a discussion of the findings, while Section VI concludes the paper and outlines future work to further enhance the robot's capabilities.

\section{Mechanism}
\subsection{Anatomy and Biomechanics of the Shoulder}
\label{sec:biomechanics}

\begin{figure}[t]
      \centering
      \includegraphics{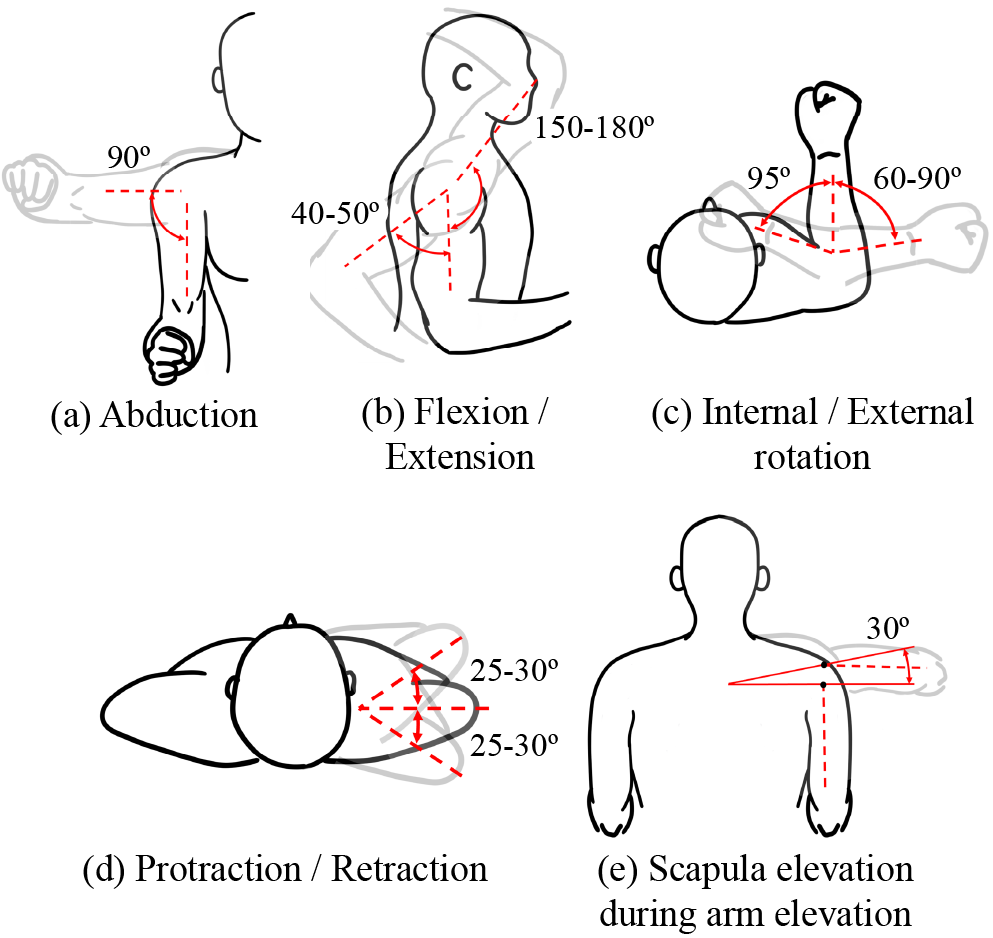}
      \caption{Shoulder normal range of motions}
      \label{fig:shoulderROM}
   \end{figure}

The shoulder's remarkable mobility and complex stabilization mechanisms, particularly the scapulohumeral rhythm, are crucial for understanding shoulder rehabilitation. The glenohumeral joint, commonly known as the shoulder, has the greatest ROM of any joint in the human body, with its normal ROM shown in Fig. \ref{fig:shoulderROM} \cite{c21,c22,c23}. This remarkable mobility is achieved by its unique ball-and-socket structure, with a shallow glenoid fossa and large humeral head allowing for multidirectional rotation with wide ROM at the cost of inherent stability. To counteract this instability, the shoulder joint incorporates various stabilizing features, with the scapulohumeral rhythm being a key component \cite{c24, c25}.

\begin{figure}[t]
      \centering
      \includegraphics{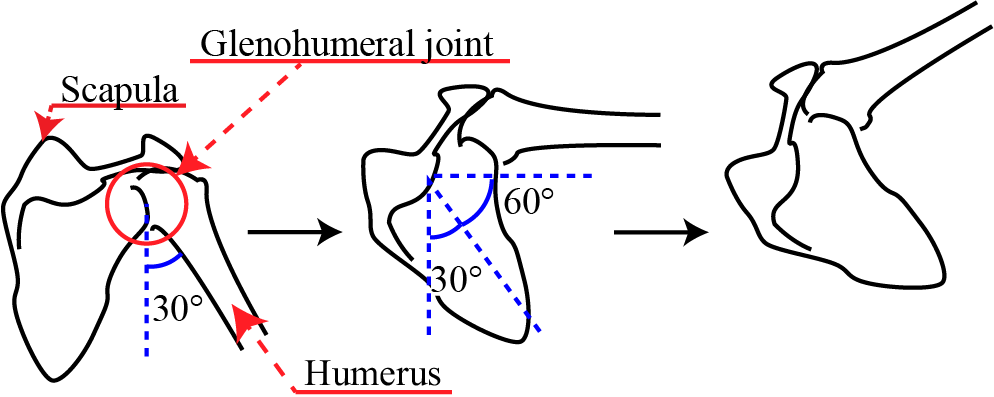}
      \caption{Shoulder anatomy with scapulohumeral rhythm during abduction}
      \label{fig:anatomy}
   \end{figure}

Scapulohumeral rhythm is an active and coordinated motion between the arm and scapula, during arm elevation (Fig. \ref{fig:anatomy}). It prevents the humeral head from becoming dislocated, avoid impingement, and minimize muscle fatigue by distributing forces across multiple muscle groups \cite{c5,c25}. Although scapulohumeral rhythm is often simplified as a 2:1 ratio of humerus elevation to scapula upward rotation, it is rather a nonlinear active movement and also can be influenced by factors such as the weight of the object being lifted and the speed of arm elevation \cite{c26, c27}. When implementing a rehabilitation program that incorporates scapulohumeral rhythm, it is essential to identify and customize the rhythm to each patient, rather than employing a universal approach \cite{c28}.

\subsection{Frozen Shoulder and the Importance of Scapulohumeral Rhythm in Rehabilitation}
\label{sec:scapulo}

\begin{figure}[t]
      \centering
      \includegraphics{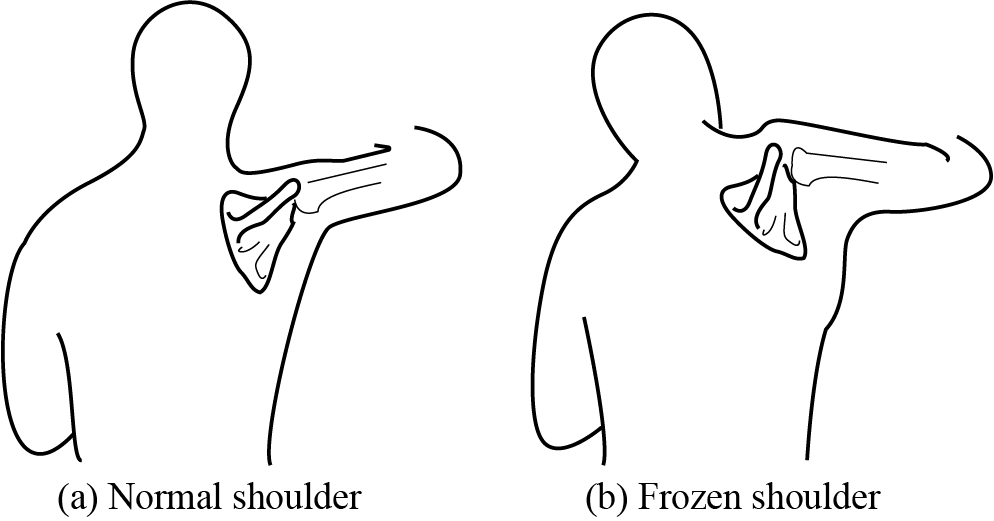}
      \caption{Compensatory shoulder shrugging in frozen shoulder during abduction}
      \label{fig:limited_shoulder}
   \end{figure}

Compensatory movements in frozen shoulder patients significantly disrupt normal scapulohumeral rhythm, necessitating rehabilitation approaches that focus on restoring proper movement patterns. Frozen shoulder is a condition that occurs when the connective tissue surrounding the shoulder joint becomes thickened and tight, resulting in a significant reduction in shoulder mobility. This limitation forces patients to adopt abnormal motor patterns, primarily appearing as trunk leaning/rotation or shoulder shrugging during arm elevation (Fig. \ref{fig:limited_shoulder}-(b)) \cite{c30, c31}. These compensatory movements can lead to overuse of the joint or muscles, causing various problems such as muscle imbalances, reduced flexibility and ROM, delayed healing, increased risk of chronic pain, and decreased overall function \cite{c29}.

Since scapulohumeral rhythm is a patterned active motion, it is essential to understand the correct movement patterns and train for proper movement patterns repeatedly during rehabilitation \cite{c32}. Rehabilitation robots designed to guide patients through appropriate ROM exercise can help patients relearn correct movement patterns, which can more effectively treat shoulder dysfunction and optimize patient outcomes.

\subsection{ROM exercise Strategy for Robot}
\label{sec:Concept of the Robot}

The proposed rehabilitation robot focuses on PROM exercise to increase ROM while maintaining normal scapulohumeral rhythm and reducing compensatory movements. To achieve the goal, the robot's ROM exercise mechanism should include:

\begin{itemize}
\item The ability to perform PROM exercise and to generate natural shoulder motion during abduction, flexion, and external rotation, with the neutral posture being the elbow bent forward.
\item The ability to correct compensatory movements by using data from the patient's unaffected side for stabilizing scapula and rigid structures for restricting trunk.
\end{itemize}

Natural shoulder motion could be achieved by incorporating two primary translation movements (protraction/retraction and elevation) and three rotational axes. These 5 movements are capable of producing abduction, flexion, and external rotation, which are the primary motions targeted in our rehabilitation approach as they represent the most common limitations in range of motion for frozen shoulder patients \cite{c37}.

Our rehabilitation exercises are designed to be performed with 90 degrees elbow flexion, which is important for isolating specific shoulder movements, particularly external rotation and abduction. When the elbow is flexed, it becomes possible to isolate pure shoulder external rotation, avoiding the forearm rotation that often occurs during this movement with a straight arm \cite{c34}. Similarly, for abduction exercises, a flexed elbow allows for focused training of the shoulder joint alone, while abduction with a straight arm, typically involves a combination of abduction and external rotation at the glenohumeral joint.

Compensatory movement control, a key feature of our design, focuses on two main areas: trunk movement and scapular elevation. Trunk compensations can be managed by immobilization, since normal shoulder motion doesn't involve trunk movements. The more complex task of controlling scapular elevation shown as shoulder shrugging is addressed through a mechanism that applies downward pressure on acromion to stabilize scapular. To ensure this pressure remains within one's normal scapulohumeral rhythm, the data from the patient's unaffected side is utilized. As bilateral frozen shoulder affects about 6-17\% of frozen shoulder patients, the majority of patients can use their unaffected side as a baseline for treatment \cite{c33}.

To achieve our ROM exercise strategy, the procedure for rehabilitation using the proposed robot system is followed:

\begin{enumerate}

\item The robot is adjusted to ensure proper coordination with the user, and the unaffected side arm is attached to the robot. On the other side, the structure beneath the armpit will be adjusted to prevent patients from leaning and rotating by limiting their trunk motion. 
\item The PROM of the unaffected side is recorded with the robot by rotating the user's arm in three main motions: abduction, flexion, and external rotation from the neutral posture. During this phase, all continuous joint angle data are recorded as the user's normal shoulder movement pattern. After reaching one's ROM, the robot holds the position for 10 seconds before returning to the starting position for ROM exercise purpose.
\item The robot is reconfigured to the other side, and the affected side is attached.
\item The robot follows the reference motion recorded in step 2. The user can set the range of exercise to gradually increase their PROM. During each motion, the robot holds the position for 10 seconds before returning, ensuring sufficient time for the exercise purpose. Additionally, the robot presses down on the user's shoulder based on their normal scapulohumeral rhythm. If the user attempts to elevate the shoulder beyond the recorded motion (shrugging), the robot suppresses the motion by blocking motion.
\end{enumerate}

\begin{figure}[t]
\centering
\includegraphics{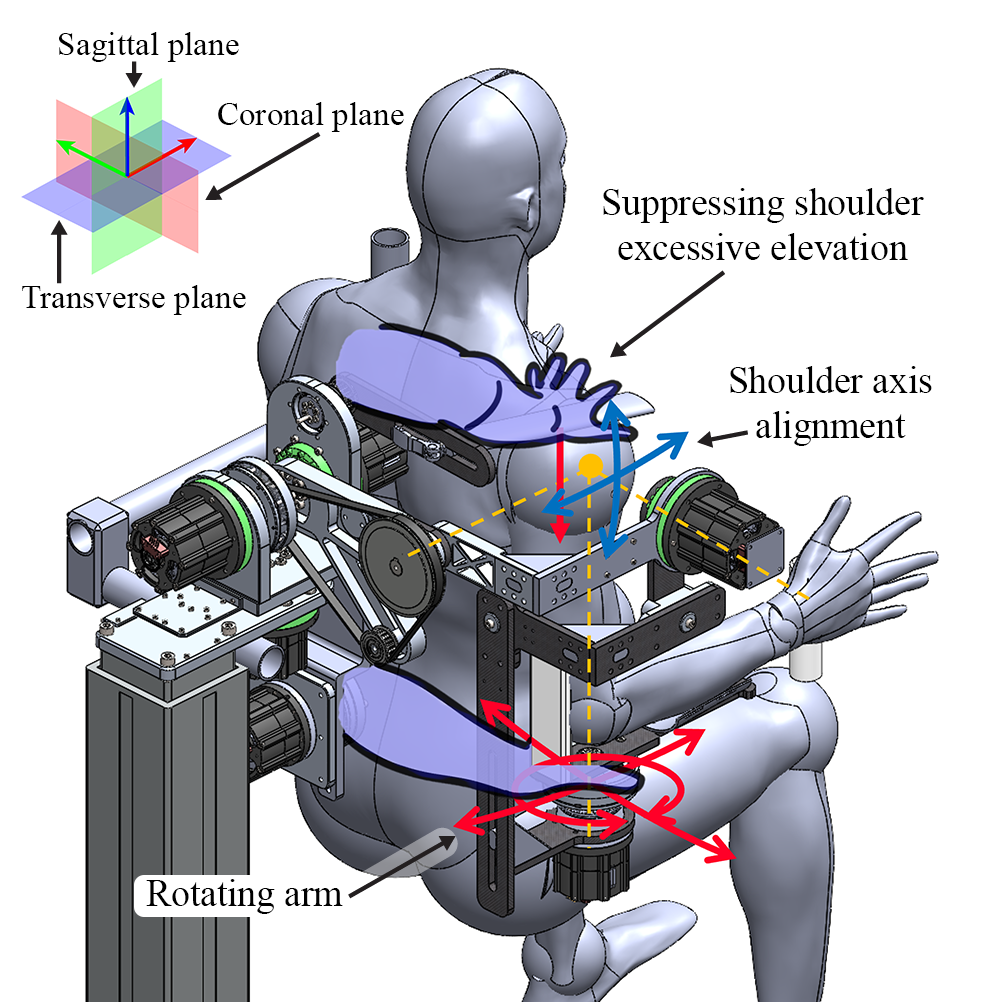}
\caption{Conceptual illustration of the robot's ROM exercise mechanism}
      \label{fig:robot_therapist}
   \end{figure}

The conceptual design of our rehabilitation robot is illustrated in Fig. \ref{fig:robot_therapist}.

\subsection{Mechanical Design for Shoulder ROM exercise}

Building upon the principles of natural shoulder motion discussed earlier, our rehabilitation robot's ROM exercise mechanism is designed with 5 DoF to replicate the complex movements of the shoulder joint. The 5 DoF mechanism incorporates two DoF for translation and three DoF for rotation. The translational movements are specifically designed to maintain proper alignment between the robot and the glenohumeral joint by mimicking scapular movements, reducing the risk of unintended joint stress or inaccurate movement patterns during exercises. The rotational movements, on the other hand, enable the robot to record shoulder rotation and perform ROM exercises in three primary directions: abduction/adduction, flexion/extension, and internal/external rotation.

\begin{figure}[t]
      \centering
      \includegraphics{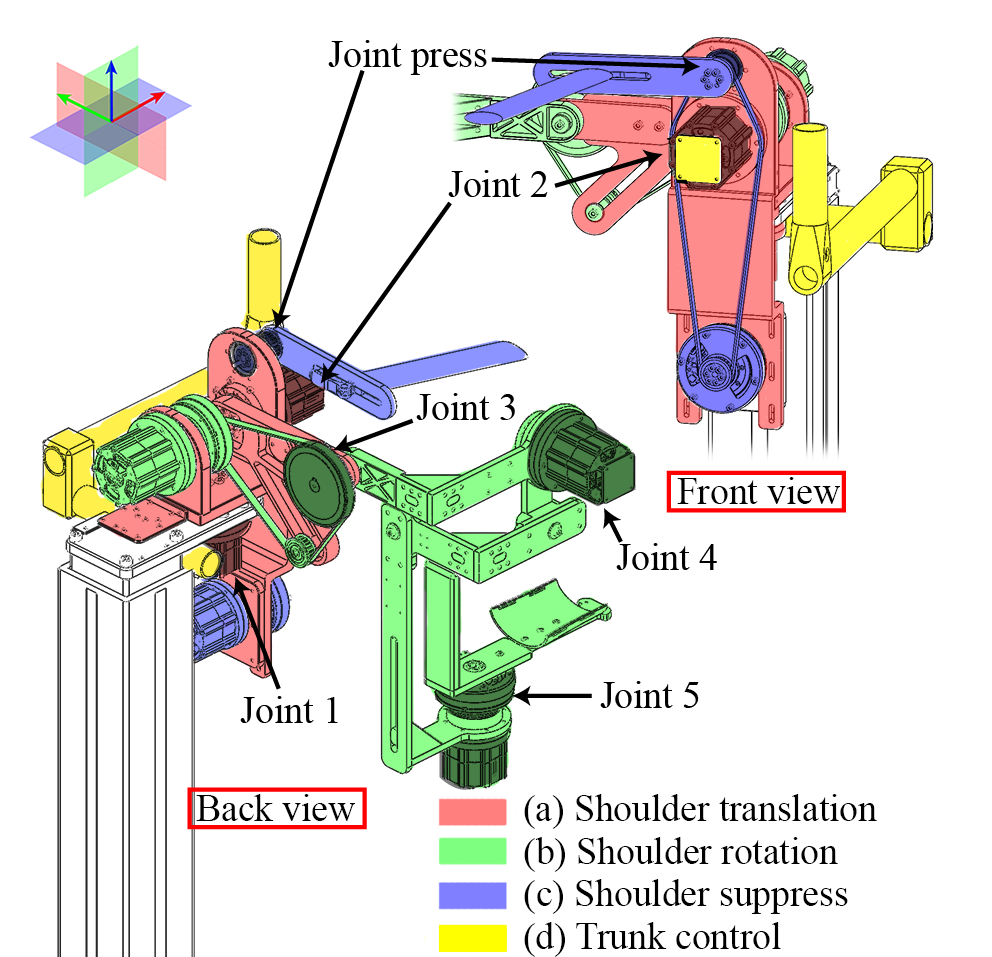}
      \caption{3D CAD model of robot with functions}
      \label{fig:3dcadmodel}
   \end{figure}

The robot's first two joints focus on shoulder translation (Fig. \ref{fig:3dcadmodel}-(a)). Joint 1, responsible for protraction/retraction, is mounted on a height-adjustable column behind the user, with its motor axis perpendicular to the ground on the sagittal plane. The Joint 2, handling scapular elevation, is positioned perpendicular to the coronal plane while remaining on the sagittal plane. This joint's axis is calibrated to align with the center of both glenohumeral joints, with its neutral position set at a horizontal level to ensure accurate and consistent measurement of scapular motion. These two joints move passively during the recording normal motion phase to follow the overall shoulder translation motion observed during shoulder's rotational movements, allowing the estimation of the shoulder’s protraction, retraction, and elevation. Together, these joints enable alignment of the robot's rotational axes with the patient’s anatomy.

Shoulder rotation is managed by the remaining three joints (Fig. \ref{fig:3dcadmodel}-(b)). Joint 3 controls abduction/adduction and is positioned behind the patient's glenohumeral joint, with its axis perpendicular to the coronal plane. Joint 4 manages flexion/extension and is positioned on the outer side of the glenohumeral joint, with its axis perpendicular to the sagittal plane. Joint 5 handles internal/external rotation and is located under the elbow from neutral position, with its axis perpendicular to the transverse plane.

To optimize weight distribution, the Joint 3's motor is positioned as a child of the Joint 1, utilizing a belt-pulley system with a 3:1 ratio. However, this configuration involves a kinematic coupling between Joints 2 and 3: when Joint 2 rotates, it induces a rotation in Joint 3 without changing Joint 3's encoder reading. To compensate for this effect, the system implements a correction algorithm that adds one-third of Joint 2's encoder value to Joint 3's reading, ensuring accurate representation of Joint 3's true rotational position.

\begin{figure}[t]
      \centering
      \includegraphics{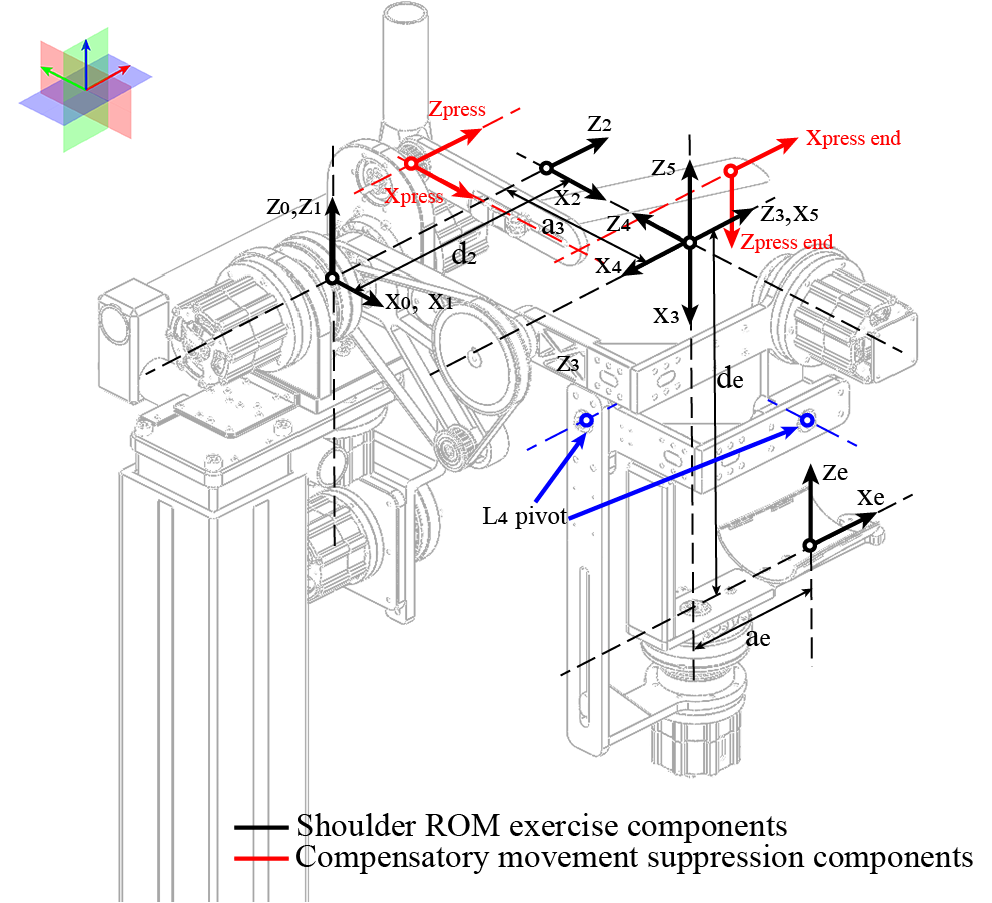}
      \caption{Denavit-Hartenberg parameter representation of robot kinematics}
      \label{fig:geometric}
   \end{figure}

To adapt various patient sizes, adjustable links are adopted between Joints 2-3 (Link 2) and 4-5 (Link 4). Each length is adjustable to accommodate shoulder width and upper arm length. Link 4's two pivot points ($L_{4 pivot}$ in Fig. \ref{fig:geometric}) allow the robot to be used on either side of the shoulder. Additionally, a medical arm immobilizer is attached on Link 5 to secure the user to the device.

\subsection{Mechanical Design for Compensatory Movement Control}

The proposed rehabilitation robot aims to suppress the compensatory movement, which includes shoulder shrugging, trunk leaning or rotating observed in patients rotating their affected shoulder.

In cases of frozen shoulder, abnormal scapular motion is primarily attributed to compensatory behaviors arising from functional changes in the surrounding muscles rather than structural deformities \cite{c16,c44}. Consequently, applying downward pressure to adjust the scapula to its normal range poses a low risk of injury. However, this process can cause discomfort due to the force applied to the shoulder. To address this, the robot incorporates a 1 DoF rotational joint (Joint press) positioned 65 mm above Joint 2 to minimize the distance between their axes while avoiding self-collision. The joint's axis is aligned along the sagittal plane, the mid-plane of the subject, to streamline alignment relative to the subject. The end-effector, designed in an elliptical prism shape, distributes pressure to minimize localized discomfort during operation. It is attached to a length-adjustable link that positions it on the acromion, enabling the recording and control of shoulder elevation for scapular stabilization.

The decision to apply downward pressure on the acromion was made because most of the scapula glides underneath the skin, making it difficult to control directly. In contrast, the acromion is a surface-level bony landmark, commonly used in motion capture systems to estimate scapular motion \cite{c56}. This mechanism suppresses excessive elevation by blocking the motion, promoting proper scapular motion throughout the rehabilitation process.

Trunk compensatory movements are managed through two stationary components. One is the angled bar positioned under the unattached arm's armpit as shown in Fig. \ref{fig:3dcadmodel}-(d). This prevents the user from leaning in the coronal plane. The other is a small square plate attached to the Joint 2 motor that the patient can lean on to, which will prevent the user from leaning in the sagittal plane.

Fig. \ref{fig:geometric} and Table \ref{table_param} show the Denavit–Hartenberg parameters and each joints' functions of the robot.

\begin{table}[h]
\caption{Modified Denavit–Hartenberg parameters of the robot}\label{table_param}
\begin{tabular}{@{}lcccccc@{}}
\toprule%
\textit{i} & $\alpha_{i-1}$ (deg) & $a_{i-1}$ & $\theta_{i}$ (deg) & $d_{i}$ & ROM (deg) & Function\\
\midrule
1 & 0 & 0 & $\theta_1$ & 0 & -30 $\sim$ 30 & Protraction / Retraction \\

2 & -90 & 0 & $\theta_2$ & $d_{2}$\textsuperscript{a} & -50 $\sim$ 0 & Depression / Elevation\\

3 & 0 & $a_{3}$\textsuperscript{b} & $\theta_3$ & 0 & -180 $\sim$ 0 & Adduction / Abduction\\

4 & -90 & 0 & $\theta_4$ & 0 & -180 $\sim$ 10 & Extension / Flexion\\

5 & -90 & 0 & $\theta_5$ & 0 & -90 $\sim$ 90 & Internal / External rotation\\

e & 0 & $a_{e}$\textsuperscript{c} & 0 & $-d_{e}$\textsuperscript{d} & - & End effector\\
\botrule
\end{tabular}
\footnotetext[a]{Corresponds to shoulder distance from Joint 1}
\footnotetext[b]{Corresponds to shoulder width.}
\footnotetext[c]{Corresponds to forearm length.}
\footnotetext[d]{Corresponds to upper arm length.}
\end{table}

\subsection{Hardware Components}

The joint motors used in the rehabilitation robot are Parker 64050 for Joint 1, Joint 5, and Joint press, while Parker 64100 is utilized for Joint 2, 3, and 4.
For the gear systems, Harmonic gear with a 50:1 ratio is employed for Joint 1-4  (CSG20-50-2UH-LW). Additionally, Harmonic gear with a 50:1 ratio is used for Joint 5  and the Joint press (CSG17-50-2UH-LW). Furthermore, Joint 3  features an additional 3:1 pulley system.
All joints are controlled using the Advanced Motion Control controllers (FM060-10-EM).

\section{Motion Planning and Control}

\subsection{Setting Neutral Posture as Zero position}

Although the robot is designed to accommodate a wide range of users, it is important to recognize that each individual's neutral posture may vary. In this study, the neutral posture is defined as the posture where the elbow is flexed at 90 degrees with the hand pointing forward and the upper arm positioned adjacent to the torso. The process of setting the zero position ensures that the robot is properly aligned with the subject's neutral posture when they are attached to the device. This zero position is then established as the baseline posture for subsequent exercises.

To set the zero position, all joints except Joints 1 and 2 are torque-controlled to move in coordination with the subject, ensuring that the robot's zero position aligns with the subject's neutral posture.

Joints 1 and 2 have their zero positions pre-set when Link 2 is leveled on both the transverse and coronal planes. During the zero position setting process, these joints are position-controlled to their pre-set zero positions, simplifying the alignment of their axes with the user's midpoint between the axes of both glenohumeral joints. This configuration also ensures proper alignment of Joint 3’s axis with the glenohumeral joint axis. The shoulder rotation joints (Joints 3, 4, and 5) are gravity- and friction-compensated, allowing them to move freely and enabling the user to comfortably adjust to their neutral posture.

As Joint Press is not physically attached to the subject, it applies a slight downward force to maintain consistent contact with the user’s shoulder. It is positioned on the acromion to follow the shoulder’s movement throughout the zero position setting procedure.

Once the user has found their neutral posture and feels comfortable, they press the user-activated button to set the zero position for all joints.

\subsection{Recording Phase}

The primary purpose of the recording phase is to capture normal shoulder movement data, including shoulder elevation, while the robot performs assisted PROM exercise on the patient. Before recording, the patient's PROM is pre-measured. This data serves as a limit during the recording process.

Once the joint for PROM exercise is selected for recording, the corresponding joint moves along a path generated by the cubic spline method from 0 degrees to the subject's pre-measured PROM at an angular velocity of approximately 5 degrees per second, rotating the upper arm while the patient remains relaxed. For abduction, Joint 3 follows this path; for flexion, Joint 4; and for external rotation, Joint 5. Among the shoulder rotation joints, those not selected for the exercise remain in their zero positions. However, during flexion and external rotation, Joint 3 lowers by the same degree as Joint 2 elevates to maintain directional consistency of movement and prevent the exercise axis from tilting due to shoulder elevation. This adjustment is necessary because if Joint 3 is locked at zero position, the abduction becomes coupled with shoulder elevation. Once the motion reaches the subject's ROM, the robot holds the position for 10 seconds before returning to the starting position.

Joint 2, which imitates shoulder elevation, is gravity and friction compensated to naturally align its rotating axis with the shoulder rotation origin. As Joint 2 can move freely, it may lower beyond the zero position, which could lead to unnecessary trunk rotation toward the attached side. To avoid this issue, Joint 2's control changes to maintain above horizontal level when it approaches or goes below the zero position.

Joint 1, designed for shoulder retraction and protraction, uses a low-gain position control strategy that mimics a spring mechanism, with the desired position set as the zero position. This control and motion planning approach addresses the issue where the patient's torso naturally returns to its neutral position after an exercise, but the robot does not. Since the torso is not rigidly fixed to the robot, such discrepancies can lead to misalignment between the robot and the patient. The spring-like behavior of Joint 1 accommodates natural shoulder movements and minor torso rotations while guiding the robot back to the zero position, ensuring proper alignment throughout the rehabilitation process.

The Joint press creates a downward force to maintain contact with the superior aspect of the shoulder, where acromion is located, allows to collect shoulder elevation data as directly as possible.

\subsection{Playback Phase}

In the playback phase, the robot utilizes the recorded encoder data from all joints during the recording phase on the unaffected side as the desired trajectory to mimic the patient's normal motion. Since the playback phase is applied to the affected side, the recorded motion is mirrored to ensure proper guidance of the affected arm, promoting symmetrical movement.

Since the playback phase involves applying the recorded motion to the affected side, safety and gradual progression are crucial. To minimize the risk of injury or discomfort, the robot incorporates a feature that allows the trajectory to be limited by a percentage of the patient's normal ROM. This adjustable limit enables the user to start with a reduced ROM and gradually increase it as their condition improves and their tolerance for the exercises grows. Once the robot reaches the user's adjusted ROM during the playback phase, it holds the position for 10 seconds before returning to the starting position, ensuring sufficient time for effective ROM exercises.

\subsection{Control}

The robot employs position and torque control to achieve its desired functionality. Position control is implemented through a conventional PID controller. This control method ensures that the robot's joints accurately follow the prescribed trajectories and maintain the desired positions throughout the rehabilitation process.

Torque control is used to compensate for gravity and friction, ensuring that the robot's movements are smooth and unaffected by external forces except the patient itself. Torque control is also employed to generate a downward force on the Joint press, allowing it to maintain constant contact with the patient's shoulder without the need for a rigid physical connection in zero position setting and recording phase.

Since the robot is designed for PROM exercises, gravity compensation must account for both the weight of the robot and the weight of the user's arm and hands, which are treated as part of the robot's end-effector as the arm is attached to the final link. The weight profiles of the upper arm, forearm, and hands were estimated using Winter's \textit{Biomechanics and Motor Control of Human Movement} (2009) \cite{c35}.

\begin{figure}[t]
  \centering
  \includegraphics{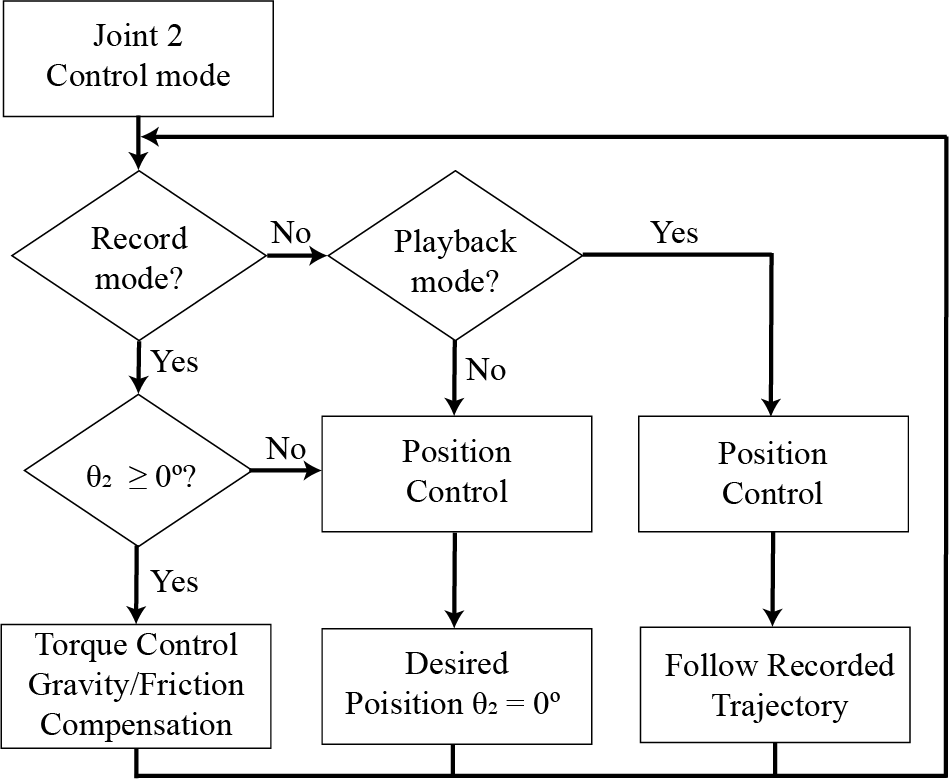}
  \caption{Joint 2's control status change by the mode}
  \label{fig:diagram}
\end{figure}

Due to the simultaneous changes in control methods during the robot's operation, Table \ref{table_control} is provided to illustrate which control method and motion planning are used in each phase and state of the robot. Notably, since Joint 2 undergoes control changes within the record mode phase, a detailed control change scheme is illustrated in Fig. \ref{fig:diagram} for further clarity.

\begin{table}[h]
\caption{Control method and motion used by phase}\label{table_control}
\begin{tabular*}{\textwidth}{@{\extracolsep\fill}lcccccc}
\toprule%
& \multicolumn{2}{@{}c@{}}{Setting zero position} & \multicolumn{2}{@{}c@{}}{Record phase} & \multicolumn{2}{@{}c@{}}{Playback phase} \\\cmidrule{2-3}\cmidrule{4-5}\cmidrule{6-7}%
Joint & Control & Motion & Control & Motion & Control & Motion \\
\midrule
1&Position& Zero position & Position\textsuperscript{a}&Zero position&Position&\makecell{Recorded\\motion}\\
2&Position& Zero position& Torque\textsuperscript{b} & \makecell{Gravity / friction\\compensated} & Position&\makecell{Recorded\\motion}\\
3&Torque & \makecell{Gravity / friction\\compensated}&Position &Zero to ROM\textsuperscript{c}&Position&\makecell{Recorded\\motion}\\
4&Torque & \makecell{Gravity / friction\\compensated}&Position&Zero to ROM\textsuperscript{d}&Position&\makecell{Recorded\\motion}\\
5&Torque & \makecell{Gravity / friction\\compensated}&Position&Zero to ROM\textsuperscript{d}&Position& \makecell{Recorded\\motion}\\
\botrule
\end{tabular*}
\footnotetext[a]{Position control with low P-gain.}
\footnotetext[b]{Position controlled below horizontal}
\footnotetext[c]{Move downward same amount as Joint 2 move upward to remain horizontal when other joint is selected to record}
\footnotetext[d]{Remains zero position while other joint is selected to record}
\end{table}

The rehabilitation robot incorporates three safety features to protect users. A user-activated button stops motion during the recording and playback phases when the user anticipates unsustainable discomfort. Upon activation, the robot stops the shoulder rotation joints (Joints 3, 4, and 5) and holds their positions for 10 seconds to perform the PROM exercise before returning to the zero position using position control. This mechanism prevents overtraining beyond the user's ROM. In addition, a software-mapped switch is included to disable the motor torque, and an emergency stop button is provided to handle critical malfunctions.

\section{Experiments \& Evaluation}

\subsection{Evaluation Key Points}

The rehabilitation robot system proposed in this study introduces an innovative feature to record the patient’s unaffected shoulder movement, including shoulder elevation, as a reference for normal motion and replicate this trajectory on the affected side to perform PROM exercises. This approach aims to suppress excessive motion caused by compensatory behaviors and stabilize the scapula commonly observed in frozen shoulder patients. To evaluate the feasibility and effectiveness of this feature, the experiments focused on five key objectives:

\begin{itemize}
\item Comparison with traditional PROM measurement: Comparison between the robot's PROM measurements and those obtained through standard clinical methods.
\item Unaffected shoulder data recording: The object is to evaluate the consistency and repeatability of the recorded normal shoulder motion data across multiple trials. This assessment is intended to determine whether the recorded data have the reliability required to serve as a reference for rehabilitation exercises.
\item Playback control performance: Assess the robot's ability to accurately reproduce recorded movement patterns using its playback algorithm.
\item Differentiation between affected and unaffected shoulders: The data acquired by the robot serves as a reference to evaluate the effectiveness of the rehabilitation approach by determining whether the guided shoulder data aligns more closely with the normal or simulated affected shoulder data.
\item Feasibility of unaffected shoulder motion application and a unique suppression system in rehabilitation: This study evaluates the ability of the affected arm to follow the trajectory of movement derived from the unaffected shoulder, up to the point of limitation of the patient's ROM, to determine the potential of this approach in rehabilitation. Specifically, the experiment investigates whether using unaffected shoulder data as a guide, combined with the suppression feature, helps the affected shoulder maintain normal movement patterns without compensatory behaviors. The results aim to assess the system's ability to stabilize the scapula within a normal scapulohumeral rhythm during PROM exercises.
\end{itemize}

For this evaluation, a healthy 35-year-old male subject (179.8 cm, 82.5 kg) with no history of shoulder disorders volunteered to participate in the experiment.

The following conventions for joint rotations are adopted in Table. \ref{table_postive}. These conventions are applied throughout the subsequent data analysis and discussion, for a clear understanding of the robot's movements and the patient's shoulder kinematics.

\begin{table}[h]
\caption{Modified joint rotation signs for analysis}\label{table_postive}
\begin{tabular}{@{}ccc@{}}
\toprule%
Joint & Positive Value  & Negative Value \\
\midrule
1 & Protraction & Retraction \\

2 & Elevation & Depression \\

3 & Abduction & Adduction \\

4 & Flexion & Extension \\

5 & External Rotation & Internal Rotation \\

Press & Upward & Downward \\
\botrule
\end{tabular}
\end{table}

\begin{figure*}[t]
  \centering
  \includegraphics[width=1.0\linewidth]{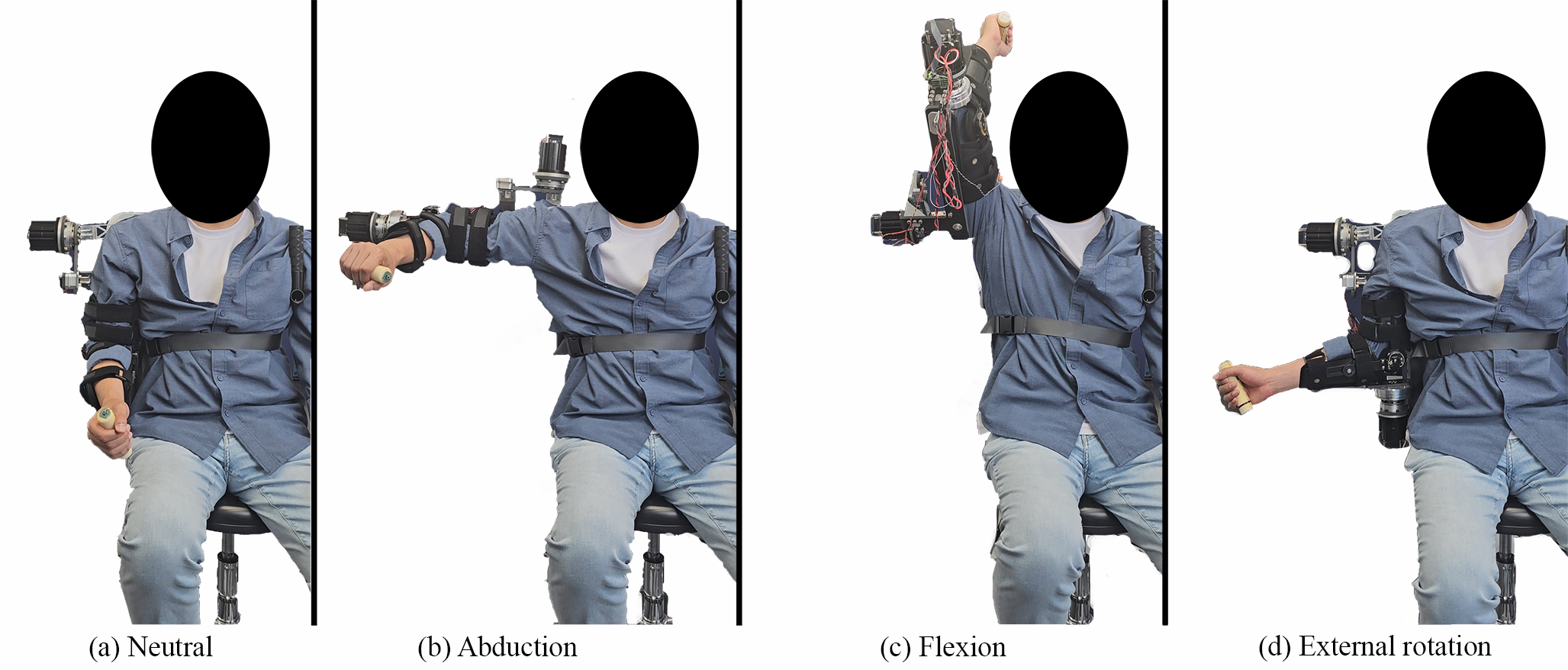}
  \caption{Exercises with proposed rehabilitation robot}
  \label{fig:exepriment}
\end{figure*}

  \begin{figure*}[t]
      \centering
      \includegraphics[width=1.0\linewidth]{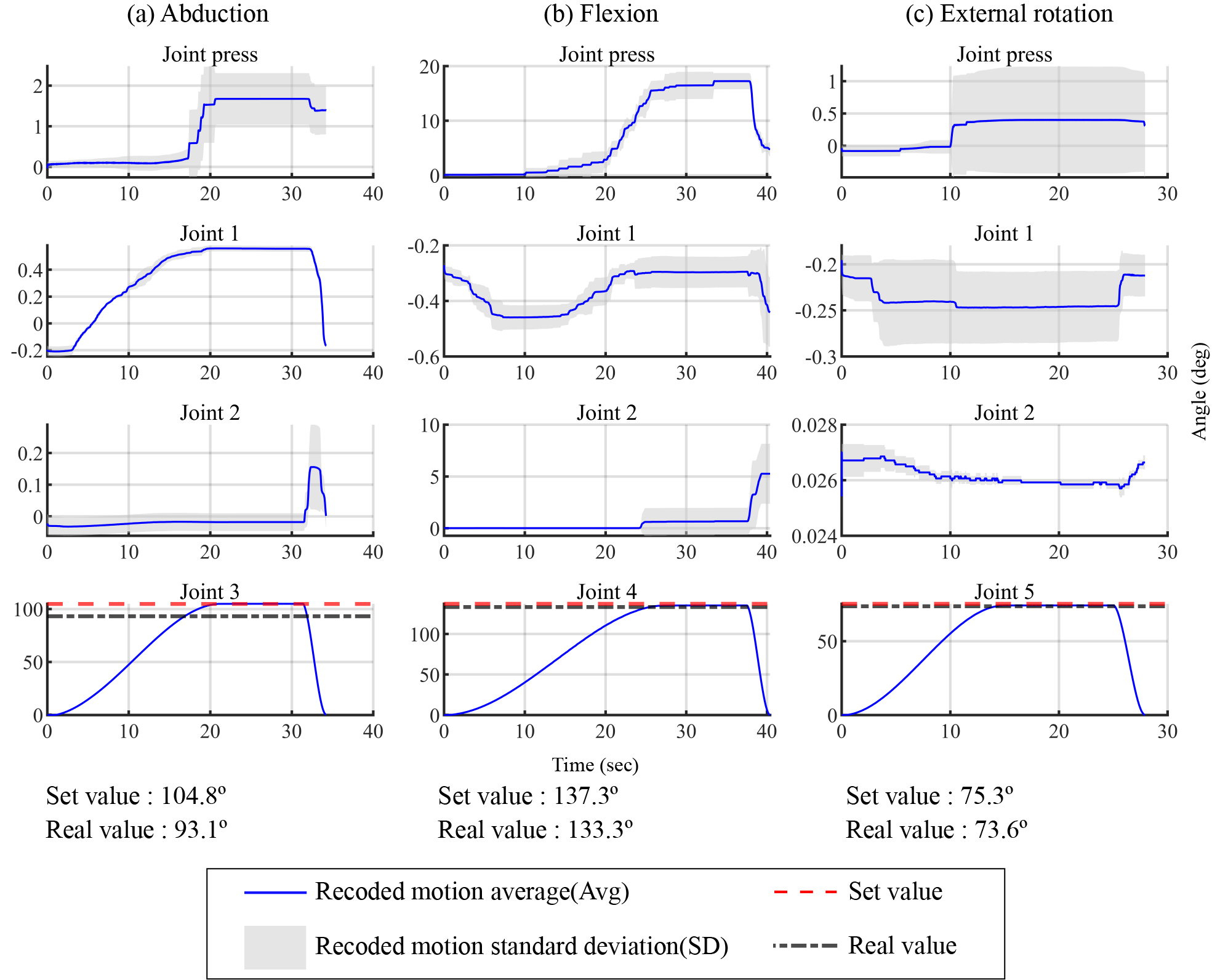}
      \caption{Motion patterns recorded on the unaffected arm: Comparison of traditional passive range of motion measurements by goniometer (set values) with robot-executed movements. The robot was programmed to target the set values, and actual achieved positions (real values) were measured with a goniometer. The figure also illustrates the repeatability of the recording process across five trials, with the line representing the average and the shaded area indicating the standard deviation.}
      \label{fig:graph_1}
   \end{figure*} 

\subsection{Comparative Analysis of Manual and Robot-Assisted ROM Assessment}

The PROM using standardized clinical measurement were measured to establish reference data and joint limits for the robotic system. Five measurements were obtained by using a goniometer for each of the following movements: abduction, flexion, and external rotation. The average of these measurements was calculated and were programmed into the robot as \textit{set values}, representing the desired end positions for ROM exercise the subject's shoulder in each specified direction.

Based upon set values, the recording unaffected arm experimental protocol proceeded as follows:
\begin{enumerate}
\item By using record mode, five repetitions were performed for each movement direction (abduction, flexion, and external rotation) as shown in Fig. \ref{fig:exepriment}.
\item Upon the robot reaching the set values, the robot's rotation joints maintain a stationary position for 10 seconds. The subject's actual achieved ROM was reassessed using a goniometer.
\end{enumerate}

The average of these secondary measurements using a goniometer is referred to as the \textit{real values} in this study.

From the robot's perspective, all three motions (abduction, flexion, and external rotation) were able to reach the set values as shown in Fig. \ref{fig:graph_1}. However, the angles measured with a goniometer when using the robot (real value) fell short of the set values that served as references.

Two potential reasons for this discrepancy between the set values and the real values can be hypothesized:

\begin{itemize}
\item Difference in reference point and dimensional translation: A discrepancy may exist between the reference points used during manual measurement and those used by the robot. This difference could arise from the challenge of translating the complex 3D movements of the human shoulder to the 2D plane in which the robot operates. This dimensional reduction could lead to systematic differences in the measured angles.
\item Imperfect alignment of rotation axes: The robot's rotation axes may not have been perfectly aligned with the subject's shoulder rotation axes. This misalignment could result in differences between the intended and actual motion.
\end{itemize}

Among these two factors, the latter may explain why the difference was more pronounced in abduction compared to the other two motions. A more detailed analysis of these observations and their implications will be presented in the subsequent subsection.

\subsection{Recorded Pattern from Unaffected Shoulder}

By analyzing the data recorded from the unaffected shoulder, distinct patterns of movement across various shoulder motions were able to be identified, providing insights into normal shoulder biomechanics and the performance of our rehabilitation device.

From Fig. \ref{fig:graph_1}-(a), the abduction data revealed several notable patterns:

\begin{itemize}
    \item Joint press: The joint press, designed to measure shoulder elevation through contact, showed minimal change, remaining close to 0 degrees up to approximately 95 degrees of shoulder abduction. However, between 95 degrees and the maximum abduction of 104.8 degrees, an average upward shift of approximately 1.67 degrees was observed. This pattern suggests that scapular upward rotation may manifest as external elevation starting around 90 degrees of abduction.
    \item Joint 1: Joint 1 exhibited a highly consistent and repetitive pattern of protraction during shoulder abduction. However, the angle of movement was less than 1 degree, indicating that the observed change was negligible.
    \item Joint 2: Joint 2, designed with gravity- and friction-compensation to maintain alignment with the glenohumeral joint axis, was expected to exhibit distinct elevation during the abduction motion. However, contrary to expectations, it remained nearly static, with changes close to 0 degrees throughout the motion. Despite its minimal movement, a slight upward spike was observed during the return phase from abduction to the neutral position, representing a repeated motion that, although small, was contrary to expectations. This behavior likely origin from the inability of Joint 2 and Joint 3 to fully replicate the natural abduction movement, resulting in a greater discrepancy between the set values and real values compared to the other two motions. Additionally, the abduction angle measured by the goniometer appears to be a composite of shoulder elevation and shoulder abduction rotation, further contributing to the observed differences between goniometer readings and robot-measured values.
\end{itemize}

With an unaffected shoulder, when the arm is bent, there appears to be minimal shoulder elevation throughout the ROM during abduction.

From Fig. \ref{fig:graph_1}-(b), several notable patterns were observed during flexion :

\begin{itemize}
    \item Joint press: Shoulder elevation was not observed until approximately 40 degrees of flexion. From that point onward, a gradual increase in elevation was noted, reaching around 2.8 degrees by 110 degrees of flexion. Toward the end of the ROM, a significant increase was observed, with elevation peaking at approximately 15.6 degrees. This pattern indicates notable scapular involvement as the flexion angle approaches its maximum.
    \item Joint 1: Joint 1 initially displayed slight retraction during the early phase of flexion, followed by protraction near 75 degrees of flexion. However, the magnitude of these movements remained below 1 degree, indicating that the changes were negligible.
    \item Joint 2: Similar to the behavior observed during abduction, Joint 2 showed minimal elevation throughout the flexion movement, in contrast to the significant elevation noted with the Joint press. Near the ROM limit, Joint 2 exhibited a slight increase in elevation. However, the standard deviation indicates that this increase cannot be considered a consistent or standardized pattern. Additionally, during the return to the zero position, a sudden elevation averaging 5 degrees was observed, mirroring the behavior noted during abduction.
\end{itemize}

Shoulder elevation was more pronounced during flexion, compared to abduction, revealing a notable relationship.

During external rotation, the Joint press showed movement around 55.2 degrees of external rotation. However, based on the standard deviation, this movement appears to result from changes in the user's posture rather than a consistent scapular motion pattern. Meanwhile, Joints 1 and 2 remained steady throughout the motion. This behavior aligns with the understanding that external rotation involves minimal scapular motion, resulting in the observed trends.

The unexpected behavior of Joint 2, particularly during arm elevation movements, can potentially be attributed to limitations in the robot's structural design. This may be explained by three factors related to the robot's configuration :

\begin{itemize}
    \item Human body flexibility and redundancy: The shoulder complex is highly flexible, with kinematic redundancy and multiple degrees of freedom. While gravity- and friction-compensation are implemented, they were insufficient to accurately represent shoulder movements driven solely by the force of the arm's motion. Additionally, the dynamic nature of tissues such as skin and fat surrounding the arm attachment introduces further variability, making it difficult to maintain consistent alignment between the robot's axis of rotation and the center of rotation of the glenohumeral joint.
    \item Lack of body-robot connection: Unlike Joint press, which maintains physical contact with the subject for direct interaction, Joint 2 lacks physical contact and feedback mechanisms to respond sensitively to the subject’s movements. This, combined with the absence of a fixed connection between the robot and the subject's torso or shoulder, exacerbates misalignment between the glenohumeral joint's center and the axis of Joint 3, contributing to the reduced effectiveness of Joint 2.
    \item Stick-slip phenomenon in Joint 2: The sudden motion observed in Joint 2 occurs due to the interplay between joint friction and the interaction with the human arm. This abrupt movement is particularly noticeable during the returning phase. As the robot applies downward force while lowering the subject's arm, the human arm provides a counterforce, resulting in a vertical force on Joint 2. The combination of this force and the joint's friction leads to a stick-slip behavior, as shown in Fig. \ref{fig:graph_1}-(a),  (b). The absence of a similar phenomenon in the opposite direction is likely due to the joint's control design, which prevents Joint 2 from moving below 0 degrees. This design feature may explain the lack of stick-slip behavior in the opposite motion observed during early elevation stage in the graph.
\end{itemize}

These three factors suggest that gravity and friction compensation alone may not be sufficient for Joint 2 to fully achieve its intended function, highlighting the need for further refinement to address the identified limitations.

Throughout the analysis of abduction, flexion, and external rotation recorded motions, the device demonstrated consistent repeatability in recording shoulder properties. This repeatability suggests that the recorded data could serve as a reliable reference for guiding rehabilitation exercises, supporting personalized treatment plans for shoulder rehabilitation.

  \begin{figure*}[t]
      \centering
      \includegraphics[width=1.0\linewidth]{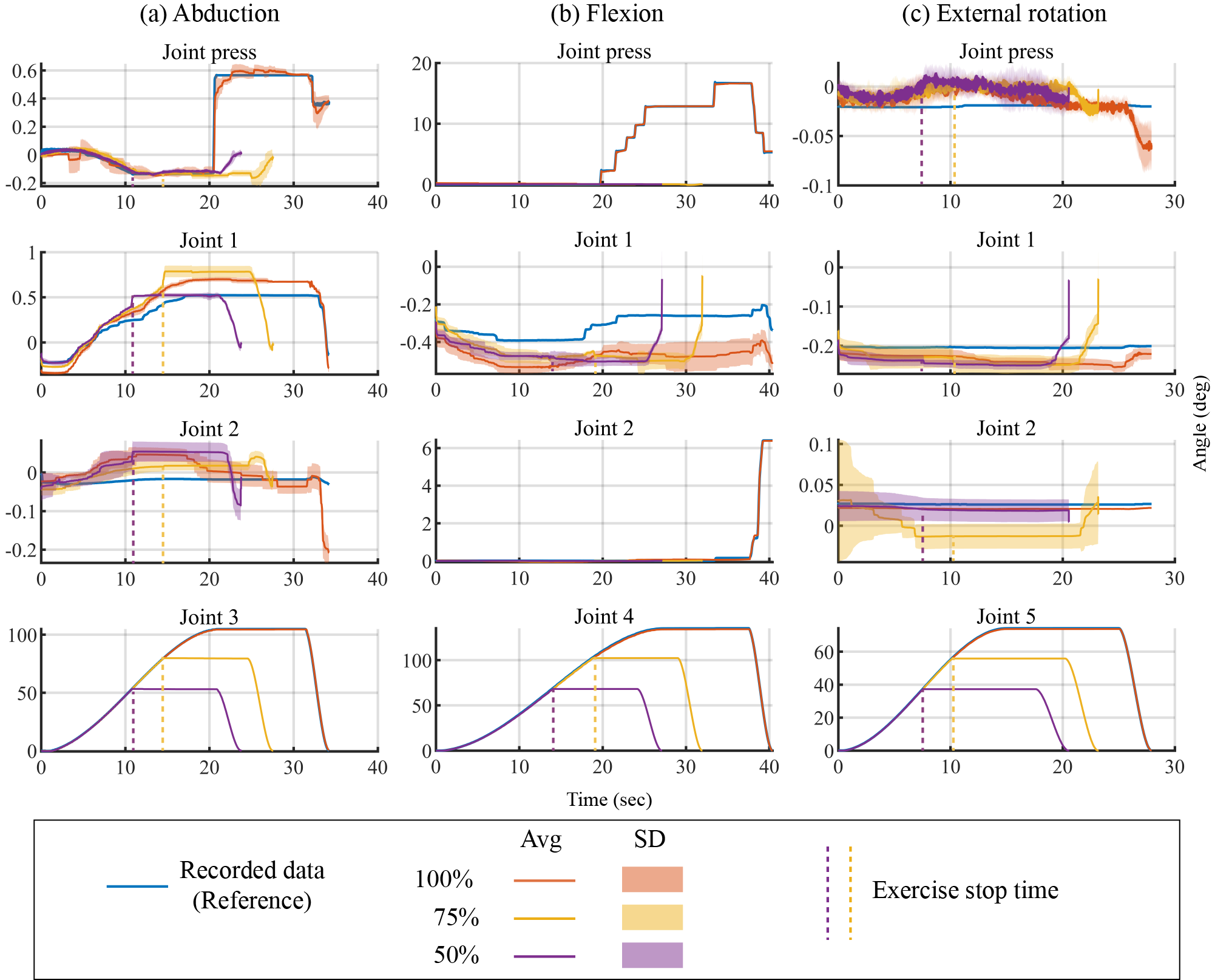}
      \caption{Comparison of playback motion with recorded reference data from the unaffected arm, demonstrating the control performance of the rehabilitation robot. The graph shows full-range playback (100\%) and partial-range playbacks (75\% and 50\% of the set value), illustrating the robot’s ability to accurately follow the reference trajectories up to the specified cutoff points and terminate motion without overshooting.}
      \label{fig:graph_2}
   \end{figure*}  

\subsection{Control Performance with Playback}

To assess the robot's ability to reproduce recorded motion patterns, playback testing was conducted under three conditions: trajectory up to the full set value, 75\%, and 50\% of the set value. One of the trials was selected from the normal recorded motion to serve as the reference trajectory. Each condition was tested five trials to ensure consistency and reliability of the results. These variations were implemented to validate the device's ability to accurately follow the reference data up to the specified cutoff points.

Fig. \ref{fig:graph_2} presents the encoder data, and the Root Mean Square Errors (RMSE), calculated between the recorded reference and the average of the full playback trials, are shown in Table \ref{table_RMSE}. The playback test demonstrated the robot's ability to closely replicate the motion patterns captured during the recording phase. Furthermore, when programmed to terminate motion at 75\% and 50\% of the set value, the device successfully stopped at the specified points without overshooting.

\begin{table}[t]
\caption{RMSE between recorded reference and average of full playback}
\label{table_RMSE}
\begin{tabular}{l c c c}
\hline
Joint & Abduction (deg) & Flexion (deg) & External Rotation (deg)\\
\hline
1 & 0.1331 & 0.1697 & 0.0314\\

2 & 0.0457 & 0.0536 & 0.0053\\

3 & 0.3527 & 0.0941 & 0.0075\\

4 & 0.0351 & 0.8906 & 0.0153\\

5 & 0.1094 & 0.0048 & 0.3444\\

Press & 0.0337 & 0.1389 & 0.0139\\
\hline
\end{tabular}
\end{table}

  \begin{figure}[t]
      \centering
      \includegraphics{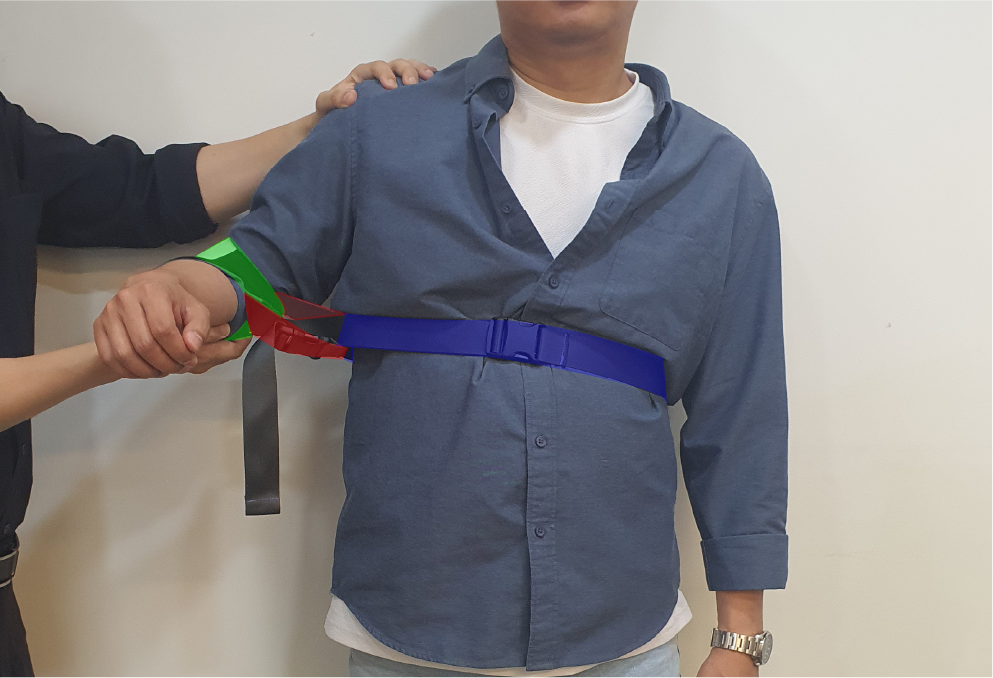}
      \caption{Simulating frozen shoulder with straps}
      \label{fig:limited_subject}
   \end{figure}
   
     \begin{figure*}[t]
      \centering
      \includegraphics[width=0.9\linewidth]{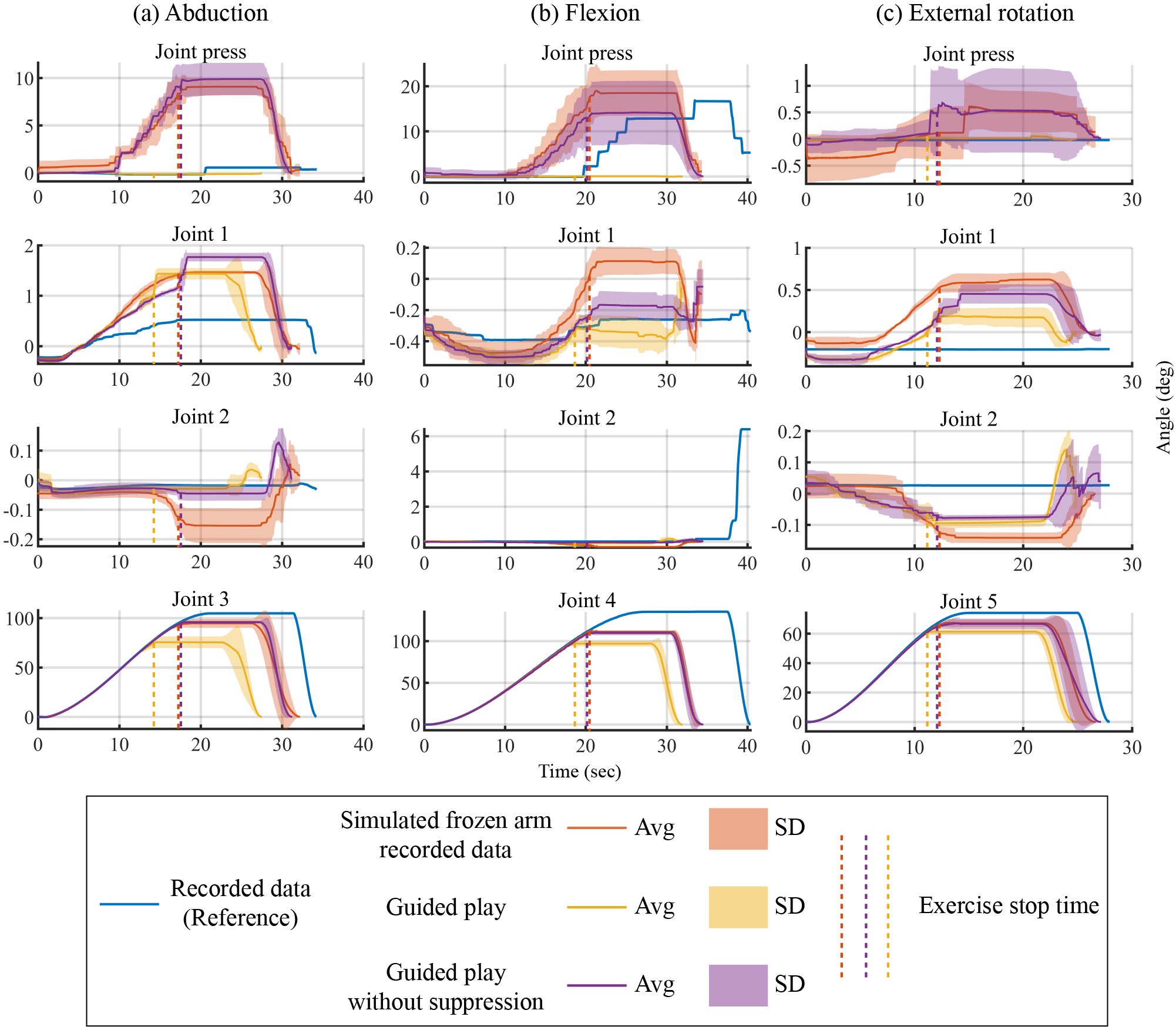}
      \caption{Comparison of simulated frozen shoulder and guided motion data (A) Comparison of motion patterns between the unaffected shoulder and simulated frozen shoulder. The earlier onset of Joint press elevation during both abduction and flexion highlights scapular dyskinesis caused by compensatory movements. (B) Comparison of guided motion (using unaffected shoulder data), simulated frozen shoulder motion, and normal motion. During guided motion with active Joint Press suppression, scapular elevation patterns more closely resembled those of the unaffected shoulder during abduction and flexion, indicating improved scapulohumeral rhythm. The earlier termination of motion during guided play underscores the Joint press's function as a pivot point, effectively reducing compensatory movements by restricting scapular elevation and emphasizing ROM exercises.}
      \label{fig:graph_3}
   \end{figure*}

\subsection{Simulated Frozen Shoulder Analysis}

This section examines the differences in movement patterns between the unaffected shoulder and the simulated frozen shoulder conditions.

To mimic the conditions of a frozen shoulder, straps were used. One strap around the torso, one strap around the elbow, and an additional strap connecting the torso and elbow straps to limit shoulder rotation as shown in Fig. \ref{fig:limited_subject}. ROM limitations achieved through this method measured with a goniometer were abduction limited to 42.4 degrees, flexion at 83.2 degrees, and external rotation at 60.3 degrees.

Five recording sessions were conducted for each shoulder rotation. The subject was instructed to press the user-activated button to halt the ROM exercise upon reaching their maximum comfortable range of motion.

The results of the simulated frozen shoulder experiment are presented in Fig. \ref{fig:graph_3}. During abduction, Joint press measurements indicated minimal elevation under normal conditions around 90 degrees of abduction. In contrast, under the simulated frozen shoulder condition, significant shoulder elevation was observed beginning at approximately 42.8 degrees of abduction, occurring about 42.2 degrees earlier than in normal motion. A 9.7-degree change in the Joint press measurements highlights pronounced shoulder shrugging compared to normal movement.

Similarly, during flexion, Joint press measurements revealed that shoulder elevation occurred significantly earlier in the simulated frozen shoulder condition compared to normal flexion, indicating compensatory shoulder shrugging. In this scenario, notable shoulder elevation began at approximately 47.2 degrees, which is 62.8 degrees earlier than in normal motion. The elevation continued to increase until exercise termination, with the Joint press recording a peak elevation averaging 20 degrees.

For external rotation, as all data ranged within 1 to 2 degrees, no notable differences were observed. This is consistent with the understanding that compensatory movements during external rotation are less associated with scapular upward rotation.

These findings demonstrate the expected data regarding the biomechanics of restricted shoulder movements. The earlier onset of shoulder elevation during abduction and flexion under restricted conditions indicates an abnormal scapulohumeral rhythm, consistent with compensatory strategies commonly observed in patients with shoulder mobility impairments. This observation might be limited by the nature of the simulated frozen shoulder and requires further investigation through clinical trials.

The final angles achieved by the robot and the PROM values measured for the simulated affected arm showed the following results: For abduction, the robot achieved a maximum angle of 95.9 degrees, while the goniometer-measured value, obtained using traditional manual PROM assessment, was 42.4 degrees, indicating a significant discrepancy. For flexion, the robot reached 109 degrees, compared to 83.2 degrees measured manually with the goniometer. In external rotation, the robot achieved 66.7 degrees, slightly higher than the goniometer-measured value of 60.3 degrees, showing relatively smaller deviations compared to arm elevation motions. The larger discrepancies observed in arm elevation motions, particularly abduction and flexion, can be attributed primarily to alignment challenges of Joint 2, as previously noted in the normal motion recording experiment, where human body flexibility, lack of body-robot connection, and frictional effects were identified as contributing factors. These differences were further amplified compared to the normal shoulder data due to the occurrence of shoulder shrugging, which increased the misalignment between the shoulder axis and Joint 2.

While the current system has limitations in achieving precise ROM measurements, the data from abduction and flexion in the restricted shoulder condition demonstrated distinct patterns compared to those of the unaffected shoulder. These results can serve as a basis for evaluating the effectiveness of the device’s unique features in the subsequent experiments.

\subsection{Application of Unaffected Shoulder Data for Guiding Limited Shoulder Rehabilitation}

The potential of the rehabilitation robot for treating frozen shoulders was assessed by applying unaffected shoulder movement data (unaffected arm data) to a shoulder with limited ROM, up to the point where the subject indicated discomfort or limitation. The playback mode was executed five times on a simulated frozen shoulder under two conditions: \textit{guided play}, where the Joint Press was active, and \textit{guided play without suppression} (hereafter referred to as \textit{without suppression}), where the Joint Press did not suppress elevation but instead recorded its movement. The results are shown in Fig. \ref{fig:graph_3}.

By analyzing the data from these two conditions, two notable observations emerge. First, in the \textit{guided play} condition, Joint Press elevation did not occur during abduction and flexion motions (associated with arm elevation), while the \textit{without suppression} condition showed shoulder elevation patterns similar to those of the affected arm. This demonstrates the potential effectiveness of the Joint Press in suppressing compensatory movements caused by scapular elevation.

Additionally, motion termination occurred earlier in the \textit{guided play} condition compared to the \textit{without suppression} condition. For abduction, termination occurred at an average of 75.5 degrees in \textit{guided play}, compared to 95.9 degrees in \textit{without suppression}. For flexion, termination occurred at 96.9 degrees in \textit{guided play}, earlier than the 109.6 degrees observed in \textit{without suppression}. In external rotation, the difference was smaller, with termination at 61.7 degrees in \textit{guided play} versus 66.5 degrees in \textit{without suppression}.

This earlier termination during abduction and flexion may be attributed to the Joint press functioning as a pivot point to suppress shoulder elevation. By restricting scapular elevation and reducing compensatory movements, the Joint press likely caused the subject to reach their ROM limit more quickly compared to other conditions. This pivot-like functionality not only emphasizes the Joint press's role in effectively restricting scapular motion but also reinforces its contribution to maintaining proper alignment and guiding motion patterns during rehabilitation exercises. This ensures the exercises remain focused on the intended ROM without allowing compensatory strategies to interfere.

\section{Discussion}

This study presents the design and development of a novel rehabilitation robot tailored for patients with frozen shoulders. By integrating a deep understanding of shoulder anatomy, biomechanics, and the unique challenges posed by a frozen shoulder, we have developed a device that addresses critical aspects of shoulder rehabilitation. Key features of our robot include:
\begin{itemize}
\item A 5 DoF mechanism that replicates natural shoulder motion, including scapular movements, to align with the glenohumeral joint throughout the ROM exercise.
\item A dedicated mechanism for managing compensatory movements, particularly shoulder shrugging, is essential for maintaining proper scapulohumeral rhythm during rehabilitation exercises.
\item A two-phase operation (recording and playback) that allows for personalized treatment by capturing the patient's unaffected shoulder movement patterns and applying them to the affected side.
\end{itemize}

Our experimental results provide strong evidence of the robot's effectiveness in guiding proper shoulder movements in the affected arm. Key findings include:
\begin{itemize}
\item PROM exercise motion data collection: The robot effectively demonstrated its ability to record PROM exercise motion data from the subject with consistent reliability and repeatability. In simulated frozen shoulder conditions, the recorded data revealed significant differences in movement patterns between the affected and unaffected shoulders, particularly during arm elevation motions. Notably, compensatory scapular elevation (shoulder shrugging) was observed to occur earlier in the affected arm compared to the unaffected arm. However, due to the performance limitations of Joint 2, accurately identifying PROM proved challenging even in the unaffected arm and was further exacerbated in the affected arm. These limitations indicate the need for further improvements to enhance the robot’s performance.

\item Guiding affected shoulder through playback performance: The robot demonstrated its ability to guide the affected arm towards normal movement patterns by applying unaffected shoulder data. With its suppressing feature as a key mechanism, the robot successfully suppressed the onset of shoulder shrugging, aligning the affected arm's motion more closely with a normal shoulder motion. This was achieved through high-fidelity playback performance, where the robot accurately reproduced recorded motion patterns, including the ability to execute partial range motions.
\end{itemize}

A unique feature of the robot, the Joint press mechanism, demonstrated its ability to suppress shoulder elevation, potentially stabilizing the scapula and guiding it toward normal movement patterns. This stabilization may help restore proper scapulohumeral rhythm, reducing stress on surrounding muscles and joints and preventing long-term complications associated with abnormal biomechanics. While shoulder elevation is closely related to scapular upward rotation, these movements differ in their properties. Advanced observation methods, such as X-ray imaging or motion capture systems, are required to further investigate this relationship and its influence on scapulohumeral rhythm.

Although the proposed robot demonstrates encouraging results, several key limitations remain that suggest clear directions for future improvement. These limitations can be categorized into two main areas: the applicability of reference data and challenges in promoting patient engagement.

A major limitation is the system’s reliance on movement data from the unaffected shoulder. This approach restricts applicability in patients with bilateral frozen shoulders or in cases where the unaffected side also exhibits abnormal kinematics. To broaden the system’s clinical utility, future research should explore the development of generalized reference datasets. Such data could be derived from population-level models or averaged normative motions, allowing treatment without depending solely on contralateral measurements.

From a usability perspective, the current system’s mechanical complexity may hinder ease of use and reduce patient compliance. As repetitive and consistent participation is essential in rehabilitation, it is critical to design systems that are more accessible and intuitive. Future design iterations should prioritize improving wearability through simplified attachment mechanisms and enhancing physical adaptability, such as automated size adjustments, ergonomic refinements, or the development of a dual-arm configuration that supports simultaneous bilateral use, thereby enhancing donning efficiency and system intuitiveness.

Moreover, effective patient engagement depends not only on mechanical simplicity but also on the provision of meaningful feedback that motivates continued participation. Although the current robot demonstrates repeatable motion patterns, the recorded joint angles show noticeable deviations compared to goniometer measurements. These discrepancies, primarily attributable to the performance of Joint 2, limit the system’s current utility as a reliable assessment tool. To address this, future designs should improve the alignment accuracy of Joint 2, enabling the robot to function as a quantitative evaluation system capable of tracking changes in the patient’s range of motion over time.

Enhanced measurement capabilities would not only provide patients with visible and motivating feedback on their progress but also support the personalization of rehabilitation protocols. By tailoring exercise regimens based on individual performance, the system could further enhance the effectiveness of long-term therapy. These considerations underscore the importance of continued research focused on improving the system’s monitoring and feedback functions.

\section{Conclusion}

This study focused on designing a robot to perform Passive Range of Motion (PROM) exercises for frozen shoulder patients while stabilizing the scapula to align with the patient’s normal scapulohumeral rhythm. The results demonstrated the robot’s ability to guide shoulder movements effectively and highlighted its potential to support proper scapulohumeral rhythm, providing a foundation for advanced rehabilitation applications. The robot’s design emphasizes accurate PROM exercises and the prevention of compensatory scapular movements, both of which are critical for improving rehabilitation outcomes.

While the initial findings are promising, further work is necessary to enhance the robot's functionality and validate its clinical utility. Key areas for improvement include refining the Joint 2 mechanism to ensure better shoulder alignment throughout the full range of motion, potentially through direct contact methods or advanced feedback sensors. Additionally, clinical trials with larger and more diverse patient populations are essential to confirm the robot's effectiveness compared to traditional rehabilitation methods and to quantify its impact on recovery times and patient outcomes.

By addressing these challenges, future iterations of the rehabilitation robot can further advance the field of robotic-assisted physical therapy, ultimately improving outcomes for patients with shoulder mobility impairments.




\section*{Declarations}

\subsection*{Ethics approval and consent to participate}
A single healthy adult voluntarily participated in this non-invasive, low-risk study. Written informed consent was obtained prior to participation, and all procedures were conducted in accordance with the ethical guidelines of the affiliated institution.

\subsection*{Consent for publication}
Not applicable.

\subsection*{Availability of data and material}

The raw encoder data (.csv) supporting the findings of this study are available from the corresponding author upon reasonable request but are not publicly accessible. For further inquiries, please contact bluerobin@bluerobin.co.kr.

\subsection*{Funding}

This work was supported in part by the Technology Innovation Program (20017345), funded by the Ministry of Trade, Industry and Energy (MOTIE), Republic of Korea.

\subsection*{Conflict of Interest}

The authors declare the following competing interests:

\begin{itemize}
\item Patent Application: Hyunbum Cho, Sungmoon Hur, Keewon Kim, and Jaeheung Park have a patent application pending (Application No. 1020230051873) with the Korean Intellectual Property Office. This patent application is related to a 'Shoulder Range of Motion Rehabilitation Robot Incorporating Scapulohumeral Rhythm for Frozen Shoulder', which is directly related to the subject matter of this manuscript.
\item Company Affiliation: Hyunbum Cho, Sungmoon Hur, and Jaeheung Park are affiliated with Blue Robin Inc., which may be involved in future development and commercialization of the technology described in this study.
\item Potential Commercialization: The results of this research may be used in future product development and commercialization by Blue Robin Inc.
\end{itemize}

These competing interests have been fully disclosed to ensure transparency. The authors affirm that these interests have not influenced the objectivity or validity of the research, its conclusions, or its reporting."

\subsection*{Authors' contributions}

Hyunbum Cho conceptualized the system, led the robot design and fabrication, designed and conducted the experiments, organized the results, and drafted the manuscript. Sungmoon Hur contributed to the robot design and fabrication, and assisted in the experimental procedures. Joowan Kim supported the experimental design and execution, and provided feedback on manuscript preparation. Keewon Kim offered medical insights and provided clinical background knowledge to support the conceptual framework. Jaeheung Park supervised the overall research process and critically reviewed the manuscript. All authors have read and approved the final version of the manuscript.

\bibliography{sn-bibliography.bib}

\end{document}